%% file: root.tex
\documentclass[letterpaper, 10 pt, journal, twoside]{IEEEtran}  %

\IEEEoverridecommandlockouts                              

\usepackage{graphicx} 
\usepackage{amsmath} 
\usepackage{amssymb}  
\usepackage[usenames, dvipsnames]{color}

\usepackage{lipsum}
\usepackage{color}
\usepackage{cite}

\usepackage{diagbox}
\usepackage{tabu}
\usepackage{balance}  

\usepackage[ruled,linesnumbered]{algorithm2e}
\usepackage{algpseudocode}
\usepackage{epsfig}
\usepackage{multirow}

\newcommand{\revision}[1]{\textcolor{black}{#1}} 

\title{
Towards Autonomous Atlas-based Ultrasound\\Acquisitions in Presence of Articulated Motion
}

\author{Zhongliang Jiang*$^{1}$, Yuan Gao*$^{1,2,3}$, Le Xie$^{2,3}$, and Nassir Navab$^{1,4}$, \textit{Fellow, IEEE} 
\thanks{Manuscript received: February, 24, 2022; Accepted April, 28, 2022.}
\thanks{This paper was recommended for publication by Editor Jessica BurgnerKahrs upon evaluation of the Associate Editor and Reviewers' comments. 
}
\thanks{$^{1}$Z. Jiang, Y. Gao, and N. Navab are with the Chair for Computer Aided Medical Procedures and Augmented Reality (CAMP), Technical University of Munich (TUM), 85748 Garching, Germany. {\tt\footnotesize{(zl.jiang@tum.de)}}
        }%
\thanks{$^{2,3}$Y. Gao and L. Xie are with the Institute of Forming Technology \& Equipment and the Institute of Medical Robotics, Shanghai Jiao Tong University, 200240 Shanghai, China.}
\thanks{$^{4}$N. Navab is with CAMP, TUM, 85748 Garching, Germany and also with the Laboratory for Computational Sensing and Robotics, Johns Hopkins University, Baltimore, MD 21218 USA.}
}

\begin{document}

\maketitle


\begin{abstract}
Robotic ultrasound (US) imaging aims at overcoming some of the limitations of free-hand US examinations, e.g. difficulty in guaranteeing intra- and inter-operator repeatability. However, due to anatomical and physiological variations between patients and relative movement of anatomical substructures, it is challenging to robustly generate optimal trajectories to examine the anatomies of interest,
in particular, when they comprise articulated joints.
To address this challenge, this paper proposes a vision-based approach allowing autonomous robotic US limb scanning. To this end, an atlas MRI template of a human arm with annotated vascular structures is used to generate trajectories and register and project them onto patients' skin surfaces for robotic US acquisition. To effectively segment and accurately reconstruct the targeted 3D vessel, we make use of spatial continuity in consecutive US frames by incorporating channel attention modules into a U-Net-type neural network. The automatic trajectory generation method is evaluated on six volunteers with various articulated joint angles. In all cases, the system can successfully acquire the planned vascular structure on volunteers' limbs. For one volunteer the MRI scan was also available, which allows the evaluation of the average radius of the scanned artery from US images, resulting in a radius estimation ($1.2\pm0.05~mm$) comparable to the MRI ground truth ($1.2\pm0.04~mm$).
\end{abstract}


\markboth{IEEE Robotics and Automation Letters. Preprint Version. Accepted April, 2022}
{Jiang \MakeLowercase{\textit{et al.}}: Towards Autonomous Atlas-based Ultrasound Acquisitions in Presence of Articulated Motion}




\bstctlcite{IEEEexample:BSTcontrol}
\input{text_final}
\revision{
\section*{ACKNOWLEDGMENT}
The authors would like to acknowledge the Editor-In-Chief, Associate Editor, and anonymous reviewers for their contributions to the improvement of this article.}

\bibliographystyle{IEEEtran}
\balance
\bibliography{IEEEabrv,references}

\end{document}

%% file: text_final.tex

\section{Introduction}

\IEEEPARstart{U}{ltrasound} (US) imaging has become one of the standard medical imaging modalities and is widely used for diagnosis and interventional objectives. US modality is powerful and effective for identifying internal lesions and organ abnormalities, particularly considering that it is non-invasive, real-time, and radiation-free.
The absence of contraindications and the potential of US imaging in vascular diagnosis suggest that it may become the main diagnostic tool to examine the extremity artery tree~\cite{favaretto2007analysis}. Particularly, US imaging is the only technique that can provide information on the degree of calcification~\cite{favaretto2007analysis}.


\begin{figure}[ht!]
\centering
\includegraphics[width=0.48\textwidth]{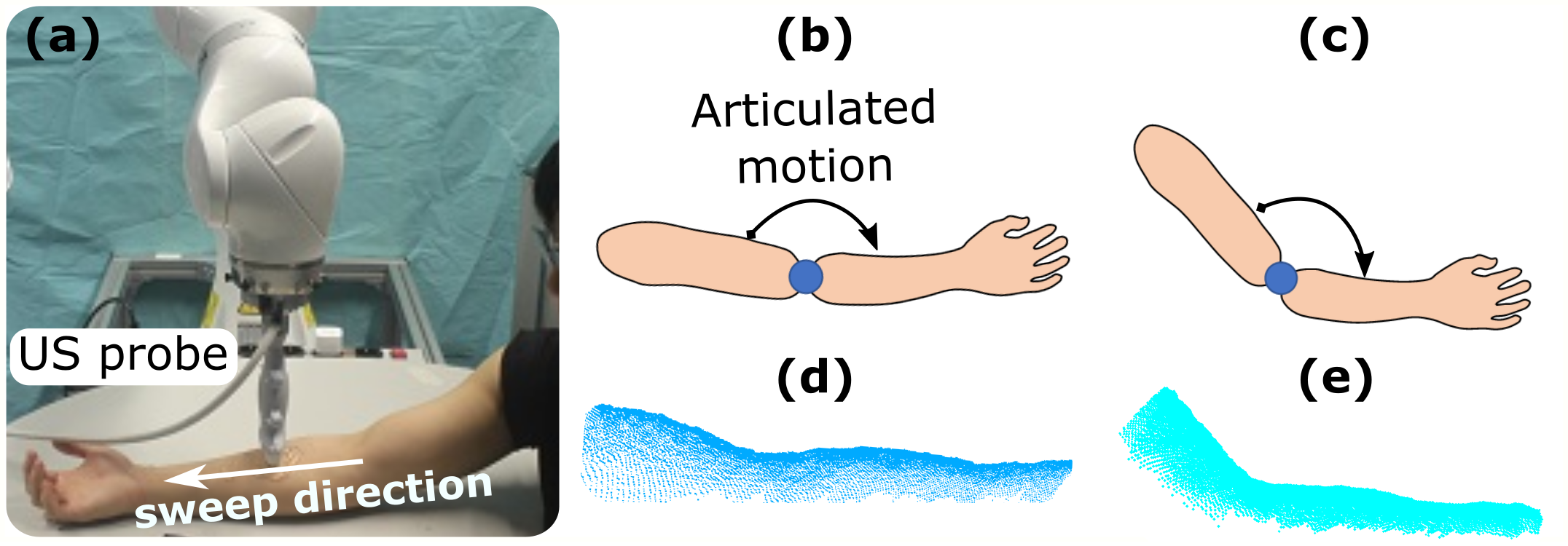}
\caption{(a) A scene of robotic US scanning of a volunteer's arm; (b) and (c) Articulated motions around the elbow joint; (d) and (e) Illustrations of the corresponding skin surfaces extracted from RGB-D data. 
}
\label{Fig_demo_question}
\end{figure}

\par

\par
To alleviate the influence caused by the inherited limitations (e.g., operator-variability) of 2D US imaging, three-dimensional (3D) US imaging has been used to characterize and quantify stenosis non-invasively for lower limb arteries~\cite{merouche2015robotic}. To obtain 3D imaging, an external spatial tracking system is often used to provide a pose for each 2D slice~\cite{morgan2018versatile}. Considering the target objects with long structures, e.g., limb artery trees, free-hand navigation of the probe will lead to distorted images due to inconsistent pressure applied between probe and objects skin. The negative effects of such distortions are exacerbated in 3D applications~\cite{jiang2021deformation}.


\par
To address these challenges, robotic US systems (RUSS) have been studied aiming to provide stable and repeatable US images. To avoid non-homogeneous deformation over the resulting 2D images, the contact force is often required to be constant during scans. To this end, Gilbertson~\emph{et al.} proposed an impedance controller for a one degree of freedom (DoF) device to stabilize US imaging during US examinations~\cite{gilbertson2015force}. Pierror~\emph{et al.} employed a hybrid force/position controller for a 7-DOF RUSS based on a 6-DOF force/torque sensor, where an external PID controller was continually run the joint position servo loop~\cite{pierrot1999hippocrate}. The implementation of such an external force controller is simple and can be easily adapted to different robotic systems. Besides the contact force, Jiang~\emph{et al.} investigated the impact of probe orientation on US imaging quality and proposed a mechanical model-based approach to accurately identify the normal direction of unknown constrained surfaces~\cite{jiang2020automaticTIE,jiang2020automatic}. Due to the intrinsic physics of US imaging, an orthogonal (normal) orientation of US probe can result in US images with higher contrast and less clutter due to a reduction of axis scattering~\cite{ihnatsenka2010ultrasound}.

\par
To achieve autonomous behavior, a feasibility study was first done by Hennersperger~\emph{et al.} by registering the 3D object surface of the current setup extracted from a depth camera to a preoperative imaging modality~\cite{hennersperger2016towards}. Then, a pre-planned trajectory on the preoperative images can be transferred to the current setup. To further consider variability between patients, Virga~\emph{et al.} employed a non-rigid registration approach to generate scan trajectories for different patients to carry out aorta screening~\cite{virga2016automatic}. Both approaches work well in their scenarios of abdominal applications. Nevertheless, the feasibility of these methods for limb arteries tree scanning will be limited due to the articulated motion as Fig.~\ref{Fig_demo_question}. Regarding limb US scans, Jiang~\emph{et al.} proposed a marker-based approach to identify patient motion; thereby, further compensating for such motions using rigid registration~\cite{jiang2021motion}. Nevertheless, the articulation around joints has not yet been investigated and is a key component of an autonomous RUSS for regular examination of the extremity artery tree.


\par
In this work, we present a vision-based approach allowing autonomous robotic US screening for limb arteries in the presence of articulated motion. To the best of the authors' knowledge, this is the first work considering the joint articulated motion for RUSS. Due to the articulated motion in real scenarios, a non-rigid registration between the arm surfaces extracted from a pre-scanned MRI and the RGB-D data of the current environment is performed. Afterward, suitable trajectories can be generated on different volunteers' arm skin surfaces. In addition, to effectively segment and track the target arteries for reconstructing 3D images of the anatomies of interest, the relative spatial information from consecutive images is exploited by incorporating channel attention modules into the standard U-Net. The proposed optical flow-based U-Net (OF-UNet) outperforms the standard U-Net in terms of dice coefficient, precision, and recall on two unseen volunteers' radial arteries. The trajectory generation approach is evaluated on six volunteers with varying poses (articulated joint angle: $120^{\circ}$, $140^{\circ}$, $160^{\circ}$). Finally, the quantitative comparison of objects' geometry (e.g., radial artery) is performed between the images obtained from robotic US scans and MRI ground truth in terms of radius. The experimental results demonstrate that comparable results are achieved in our setup (robotic US: $1.2\pm0.05~mm$, MRI: $1.2\pm0.04~mm$). The code\footnote{code: https://github.com/jiang96998/AutoLimbUSscan} and video\footnote{video: https://www.youtube.com/watch?v=OlBwmqLGf7w} can be publicly accessed.


\section{Related Work}
\subsection{Cross-Sectional US Image Segmentation}
\par
Due to the intrinsic physics of US modality, it often suffers from artifacts such as gross reverberation, contrast reduction, and clutter. These characteristics lead US imaging to become one of the most challenging modalities for automatic segmentation~\cite{mishra2018ultrasound}. Focusing on the vascular application, the Hessian matrix based vessel enhancement filters, e.g., the Frangi filter~\cite{frangi1998multiscale}, was developed in the early studies and is very lightweight allowing real-time use. But these approaches cannot provide high-accurate segmentation results. 

\par
To accurately extract vascular outlines, Karami~\emph{et al.} proposed an adaptive polar active contours approach for the internal jugular vein from video clips~\cite{karami2018adaptive}. By introducing the weighted energy functions, the local information is considered; thereby, a robust segmentation result was achieved even when imaging quality is poor. However, this method requires manual segmentation of the first frame. To allow automatic segmentation, Smistad~\emph{et al.} presented an ellipse template-based approach~\cite{smistad2015real}. Since the segmentation result is forced to be an ellipse, the vessel boundary cannot be accurately depicted, particularly for unhealthy ones. In addition, Abolmaesumi~\emph{et al.} compared the performance of five popular feature tracking approaches on the carotid artery and claimed that the Sequential Similarity Detection method and the Star-Kalman algorithm achieve the excellent performance, while the correlation and Star algorithms exhibit poorer performance with higher computational cost~\cite{abolmaesumi2002image}.

\par
The convolutional neural network (CNN) has achieved phenomenal success in various computer vision tasks and it has been successfully applied to CT and MRI segmentation tasks~\cite{litjens2017survey}. The U-Net architecture is one of the most successful representatives proposed for biomedical image segmentation~\cite{ronneberger2015u}. This network with pyramid structure and skip connections is extended from a fully CNN and has been widely used for different tasks, e.g., segmenting vascular structure~\cite{jiang2021autonomous} from B-mode images. Besides, Mishra~\emph{et al.} proposed a fully CNN with boundaries attention to extracting vessels from US liver images~\cite{mishra2018ultrasound}. Chen~\emph{et al.} developed a recurrent fully CNN architecture to capture salient image features at multiple resolution scales for peripheral vessels~\cite{chen2020deep}. They reported comparable performance to a human operator. In addition, Jiang~\emph{et al.} incorporated color Doppler images to train the VesNetSCT+ for small vessels, e.g., femoral and tibial artery~\cite{jiang2021automatic_baichuan}. 
Compared with classic approaches, learning-based approaches have demonstrated promising potential for real-time performance and unstructured segmentation results. 

\subsection{Automatic Robotic US Acquisition}
\par
The development of automatic robotic US acquisition is very meaningful to provide reproducible images and consistent diagnosis results and it is helpful to address the work-related musculoskeletal disorders of sonographers. Furthermore, since such autonomous systems can separate the patients and clinicians, they can continue providing healthcare intervention, even during infectious pandemics, e.g., COVID-19. Ma~\emph{et al.} presented a vision-based framework to automatically perform US examinations of lungs without manual localization~\cite{ma2021autonomous}. Besides, Hennersperger~\emph{et al.} validated the feasibility of autonomous US scans for abdominal application based on a preoperative MRI image~\cite{hennersperger2016towards}. The robotic probe scanning trajectory is generated by registering the object surface point cloud to the MRI surface and updating the trajectory by performing US-to-MRI registration using LC$^2$ similarity.

\par
Regarding the automatic acquisition of vascular objects, Jiang~\emph{et al.} employed the real-time US images as feedback to plan the scanning direction online~\cite{jiang2021autonomous}. The B-mode images are segmented by a U-Net and further feed to a ring buffer to optimize the local vessel centerline. Besides, Huang~\emph{et al.} imitated clinical protocols to automatically position a probe in the longitudinal direction of the carotid~\cite{huang2021towards}. Both of the aforementioned methods require an initial positioning of the probe to the position that can visualize the target vessel. In addition, various automatic acquisition approaches have been proposed for different applications, e.g., breast cancer scans~\cite{welleweerd2021out} and reinforcement learning-based standard US plane acquisition~\cite{bi2022vesnet}.

\par
Specific to the screening of limb artery trees, Jiang~\emph{et al.} presented a motion-aware system based on an RGB-D camera and passive markers to identify and compensate for the limb motion happened during scans~\cite{jiang2021motion}. However, this work assumed the motion is rigid, and it is only tested on a phantom.
In real scenarios, articulated motions around joints are inevitable, the pioneering work of MRI-based approach~\cite{hennersperger2016towards} is not applicable for this application. To automatically and properly generate trajectories for different patients in clinical, the limb articulated motion should be taken into consideration.




\begin{figure*}[ht!]
\centering
\includegraphics[width=0.85\textwidth]{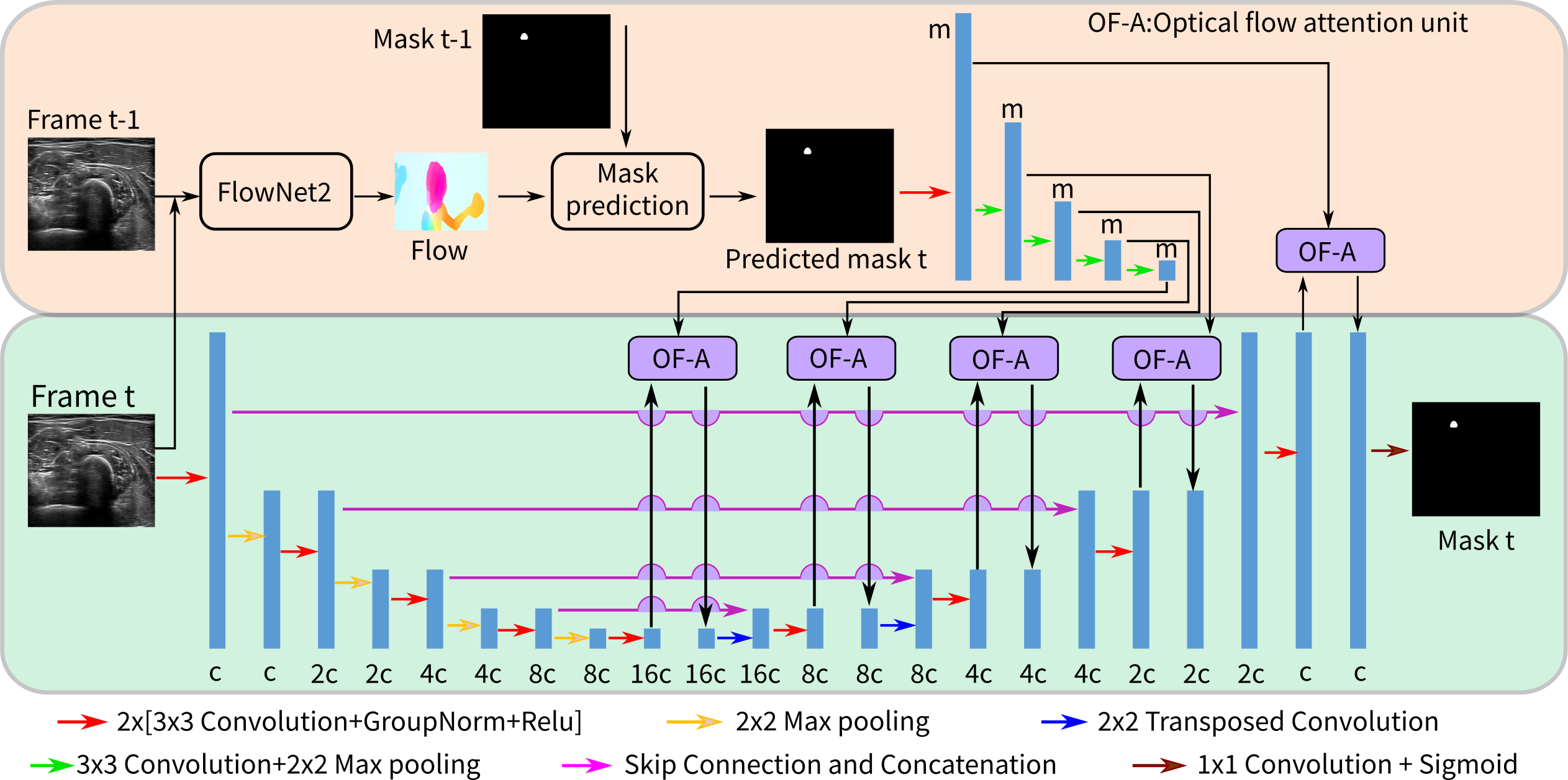}
\caption{Illustration of the proposed optical flow-based Unet architecture.}
\label{Fig_network_structure}
\end{figure*}

\section{Ultrasound Vessel Segmentation}
\subsection{Data Collection}
\par
In this work, all US images were recorded from ACUSON NX3 US machine (Siemens Healthineers, Germany) using a linear probe VF12-4 (Siemens Healthineers, Germany). Since the limb arteries are often located at shadow layers, the depth of the images was set $4.5~cm$. The other acquisition parameters were set as follows: THI: $9.4~MHz$, DB: $60~dB$, focus: $2~mm$. To save the real-time US images, a frame grabber (Epiphan video, Canada) is connected to US machine using OpenCV. The recording frequency is $30~fps$.

\par
To evaluate the proposed system on limb arteries, the radial artery is selected as the target vessel, which starts from the elbow and ends at the wrist. The upper arm part is called the brachial artery. Volunteers were asked to sit on a chair and position their arms in different poses on a flat table. Then the US probe is manually moved from the elbow towards the wrist. 
Regarding the training data, five sweeps ($6362$ frames in total) were acquired from three volunteers (Age: $28.3\pm2.5$  BMI: $23.0\pm0.2$). The ratio between training and validation is $9:1$. To further test the trained model on unseen data, $500$ images were recorded from two additional (unseen) volunteers (Age: $26.5\pm1.5$  BMI: $21.9\pm1.8$) with the same acquisitions parameters. All images ($256\times 256$ pixel) are collected and carefully annotated under the guidance of an experienced clinician.

\subsection{Network Architecture}
\par
Compared with carotid and aorta, the size (diameter) of radial artery (mean$\pm$SD) is only $2.36\pm0.41~mm$~\cite{beniwal2014size}. Due to the extremely small size and the artifacts of US images, it is very challenging to accurately segment and track limb arteries, like a radial artery. To accomplish this task, Jiang~\emph{et al.} incorporate color Doppler images as channel attention to improving the segmentation results for limb arteries. Generally, better performances have been achieved using the proposed VesNetSCT+ compared with the standard U-net. Nevertheless, the quality and stability of the real-time Doppler images are very crucial for the final performance. Considering the color Doppler images are sensitive to the probe orientation and some US machine cannot robustly provide stable and accurate color images during the scanning, this work first introduces the spatial optical flow images to assist the real-time segmentation and tracking of limb arteries. The illustration of the proposed OF-UNet is depicted in Fig.~\ref{Fig_network_structure}.

\par
The ideal of the OF-UNet is inspired by the concept of the analog signal, which is continuous, e.g., force, and temperature. Regarding US scan of an anatomy, the successive two B-mode images are continuous when the US recording frequency is high enough. In other words, the vessel's position and geometry in two successive US images are close to each other. Benefited from this knowledge, the mask segmented from the previous frame can be used to generate an attention map to facilitate the accurate segmentation of the current frame. To further consider the tissue variation caused by the probe movement in the successive US images, flownet2~\cite{ilg2017flownet} is employed. The flow images $F$ are estimated by feeding two successive frames (the current image $I_t$ and the previous image: $I_{t-1}$) to the pretrained FlowNet2. Combined with the mask of the previous frame, the potential mask of the current frame $M_{t}$ is estimated based on the current flow image $F_t$ and the mask of the previous frame $M_{t-1}$ using Eq.~(\ref{eq_predictMask}).

\begin{equation}
\label{eq_predictMask}
\begin{aligned}
&{P}^{mask}_{t-1} = \{p=(i,j)|m(i,j) == 1\}~~~m(i,j)\in M_{t-1} \\
&{P}^{mask}_{t} = \{p+F_t(p)\}~~~~~~~~~~~~~~~~~~~~\forall p\in{P}^{mask}_{t-1}\\
&M^{'}_t(i, j) =  
    \begin{cases}
    1 & \text{if} (i,j)\in P^{mask}_{t}\\
    0 & \text{others}
\end{cases} \\
\end{aligned}
\end{equation}
where $p=(i,j)$ is the pixel position, ${P}^{mask}_{t-1}$ is the set of the pixel positions, where the image value is one on $M_{t-1}$. ${P}^{mask}_{t}$ is the prediction of the mask of the current frame based on the flow image. $M^{'}_t(i, j)$ is the predicted binary mask of the current image. 

\par
To maximally utilize $M_{t-1}$ for accurate segmentation, the predicted mask $M^{'}_t$ is encoded using a $3\times3$ convolution kernel and followed by a $2\times2$ max pooling kernel. Then, five data representations are computed, which are corresponding to the data obtained in different layers of the U-net encoder. Compared with the standard U-Net, the optical flow based attention module (OF-A) is applied multiple times after each skip connection. The skip connections between encoder and decoder are used to provide more features from the shallow layers. The OF-A module is implemented channel-wise with two inputs obtained from the prediction mask $F_m$ and the concatenated feature representations of raw images $F_c$. To balance $F_m$ and $F_c$, a convolution process is first used to encode $F_c$ from $N$ channel (e.g., $16$c, $8$c, etc.) to $m$ channel. Then, the result of the convolution and $F_m$ are concatenated and it is further encoded to a one-channel representation $F_a$. The output of OF-A module $F_o$ is computed as Eq.~(\ref{eq_attention_result}).

\begin{equation}\label{eq_attention_result}
F_o = F_c \oplus [F_c\otimes \text{Sig}(F_a)]
\end{equation}
where $\oplus$ and $\otimes$ denote element-wise addition and channel-wise multiplication, respectively.

\section{Autonomous Robotic US Scanning}
\par
To automatically generate scan trajectory for robotic execution, a \revision{pre-scanned} MRI image is employed as a template. The radial artery and arm surface are carefully annotated by an expert from the MRI image using MITK\footnote{https://www.mitk.org}. Section IV-A presents the details for generating a scan trajectory on the skin surface of the atlas. The human arm surface extraction approach using an RGB-D camera in the current environment is described in Section IV-B. In Section IV-C, the non-rigid registration approach addressing the potential articulated motion and variations between patients is presented. Finally, an object servoing method is developed to horizontally center the target vessel in Section IV-D. The horizontally centering process is used to alleviate the negative impacts caused by the non-rigid registration errors. 


\subsection{Scan Trajectory Generation on Template}
\par
\begin{figure}[ht!]
\centering
\includegraphics[width=0.40\textwidth]{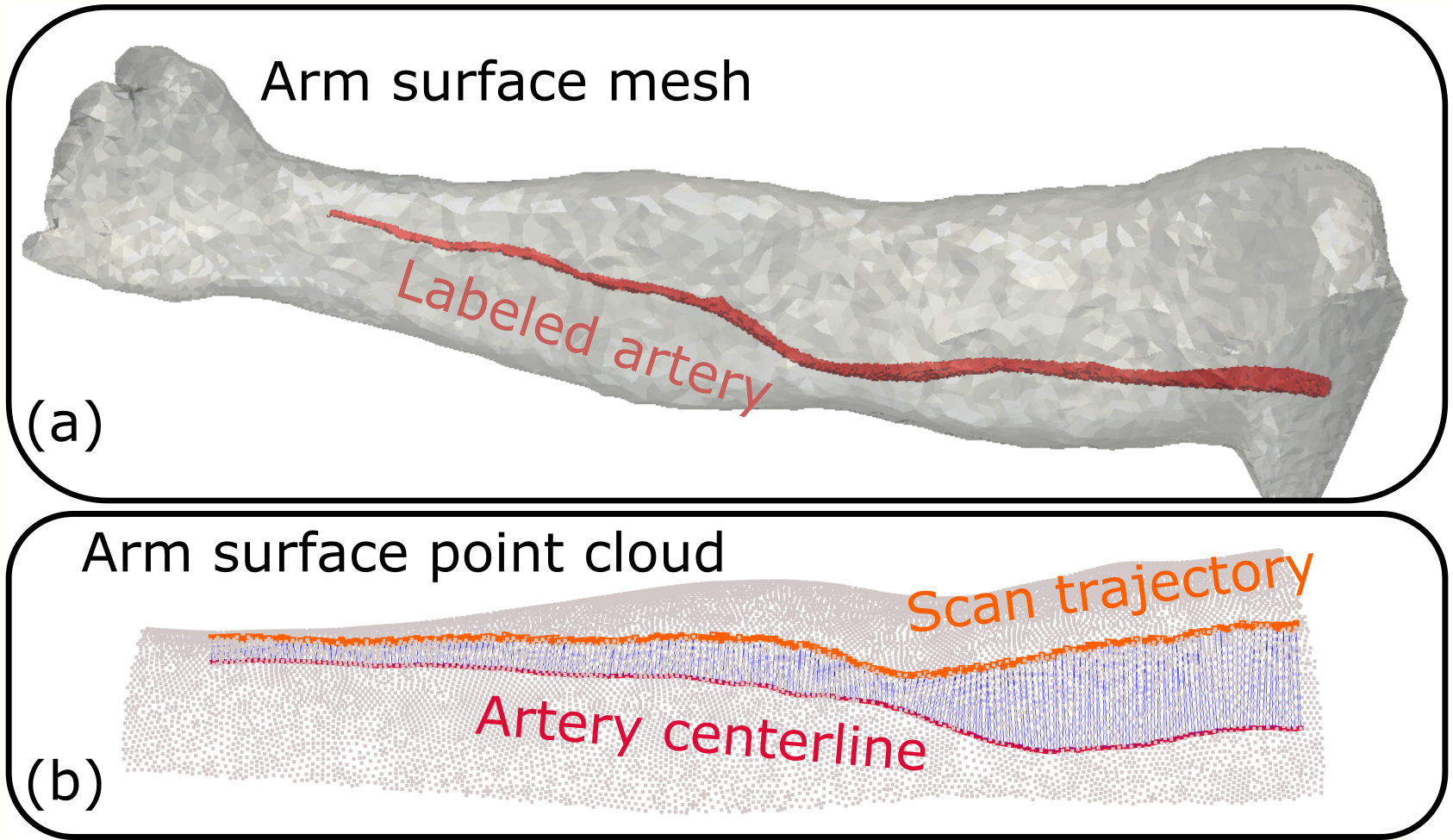}
\caption{Generation of scan trajectory on atlas template. (a) depicts the arm surface and labeled target arteries, (b) presents the generated scan trajectory on arm surface. The blue lines are the correspondences between the centerline points and the generated trajectory points.
}
\label{Fig_scan_trajectory_generation}
\end{figure}

In this study, a template MRI image of a volunteer's arm is used as a generic atlas template. The objective arteries (radial and brachial arteries) underlying the skin surface are manually annotated by an expert [see Fig.~\ref{Fig_scan_trajectory_generation}~(a)]. To generate a smooth point cloud from the mesh, the Laplacian meshes processing method is used and followed by the least squares upsampling method in MeshLab\footnote{https://www.meshlab.net}. The resulting arm surface point cloud is shown in Fig.~\ref{Fig_scan_trajectory_generation}~(b). To generate the scan trajectory on the skin surface, we first refine the centerline of the tubular artery using VMTK\footnote{https://www.vmtk.org}. Afterward, a scan trajectory on the skin surface is generated by projecting the vessel centerline upwards, which can guarantee both the visibility of objects and the contact between probe and skin. Since the arm surface point cloud is dense, the projected trajectory is generated by searching for the nearest points of individual vessel centerline points from the surface point cloud $\boldsymbol{S}$. The corresponding points among the generated trajectory and the vessel centerline have been connected using a blue line in Fig.~\ref{Fig_scan_trajectory_generation}~(b).


\subsection{Depth-based Arm Surface Extraction from RGB-D Images}
\par
\begin{figure}[ht!]
\centering
\includegraphics[width=0.48\textwidth]{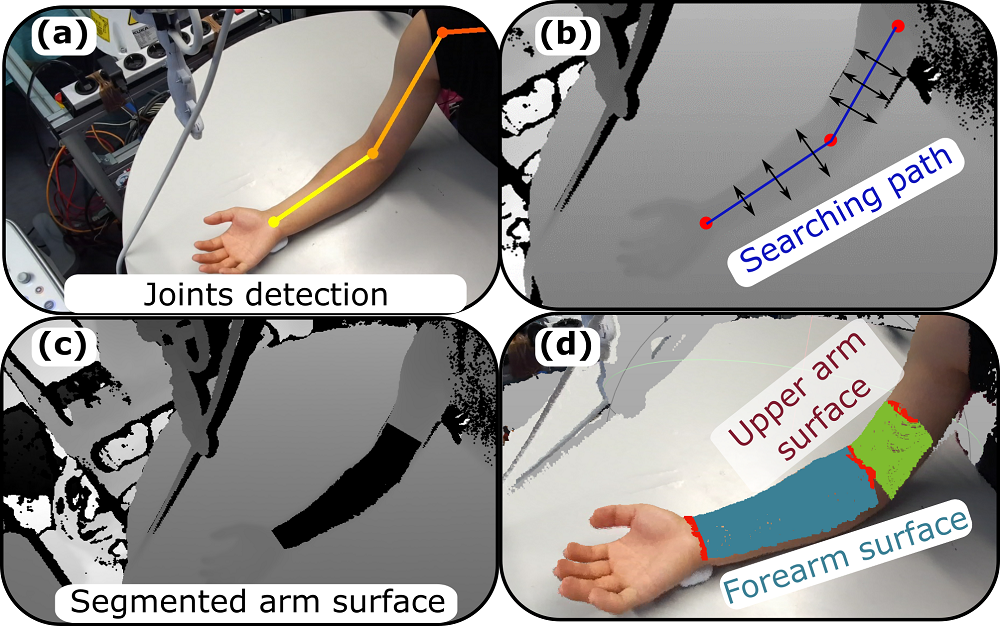}
\caption{The illustration of arm surface extraction procedures. (a) RGB image with the joints' positions detected using OpenPose. (b) depth image with identified joints. (c) segmented result annotated on a depth image. (d) colored point cloud scene with the identified forearm and upper arm point cloud.}
\label{Fig_arm_segmentation}
\end{figure}

To enable automatically screening of limb arteries, the planned path on the atlas needs to be further transferred into the current setup. To capture the current environment, an RGB-D camera (Azure Kinect, Microsoft Corporation, USA) is employed. To accurately and robustly extract the arm surface from the noisy environment, an adaptive threshold-based segmentation approach is applied to depth images. In addition, to consider the articulated motion, OpenPose\footnote{https://github.com/CMU-Perceptual-Computing-Lab/openpose} is employed to identify the joints position on the RGB image [see Fig.~\ref{Fig_arm_segmentation}~(a)]. After aligning the depth images to the RGB images, the pixel position of the key joints (wrist, elbow, and shoulder) are marked on depth images as red circles. Since the volunteer's arm was placed on the flat table, the depth of the arm surface is different from the background. Thereby, an adaptive threshold-based method is proposed to extract the arm surface. 

\par
Considering the articulated motion, the surface of the forearm and upper arm are identified separately. Regrading the forearm surface, we first connect the pixel positions of wrist and elbow joints as search path [blue line in Fig.~\ref{Fig_arm_segmentation}~(b)]. Then, a seed point is initialized at the wrist joint and a searching process is performed bidirectionally along the direction normal to the search path. The searching 
directions are marked as black arrows in Fig.~\ref{Fig_arm_segmentation}~(b). The depth difference feature $f_d$ is computed using \revision{Eq.~(\ref{eq_depth_difference})}.

\begin{equation}\label{eq_depth_difference}
\revision{
f_d(i)= I_d^{2}(P_i) - I_d^{2}(P_{i-k})
}
\end{equation}
\revision{where $I_d$ is the depth value at individual pixels in $mm$. $P_i$, and $P_{i-k}$ are two sampling points with $k$ pixel distance on the bidirectional arrow [Fig.~\ref{Fig_arm_segmentation}~(b)]}, $k=2$ in this work.

\par
Once $f_d(i)$ is larger than a preset threshold $T_{d}$, $P_i$ is considered as the arm surface boundary. Afterward, the aforementioned process is repeated at the next seed point, which is also located on the search path with a fixed distance from the last seed point. However, besides the first searching process, the second termination condition is applied based on the local distance $L_d$ computed between the seeds point and the identified edge point. Considering the human arm boundary is continuous and smooth, there should not be a large change in two consecutive $L_d$. Thereby, once the local $L_d^{i}$ is larger than the adaptive value $L_d^{i-1}+T_{l}$, the searching process will be terminated and the current location of the searching agent in the search direction is seen as the edge point. $i$ is the iterator of the current seed point. $L_d^{i-1}$ and $T_{l}$ are the distance computed at the former seed point ($i-1$) and a preset threshold, individually. Then the whole process can be repeated again for upper arm segmentation starting from elbow joint. The segmentation result has been marked both on depth images and on \revision{colorized point cloud of scenes} in Fig.~\ref{Fig_arm_segmentation}~(c) and (d), individually.



\subsection{Non-Rigid Registration} 

\subsubsection{Initial Alignment}
\par
To achieve robust registration performance, an initialization procedure is carried out to align the source point cloud $\boldsymbol{S}$ obtained from MRI and the target point cloud $\boldsymbol{T}$ obtained from RGB-D images. To align $\boldsymbol{T}$ with $\boldsymbol{S}$, the required transformation can be computed by overlaying the identified joints to the labeled wrist, elbow and shoulder joints on the atlas. To further consider the variation between different patients, the template atlas $\boldsymbol{S}$ is scaled along the three PCA principle directions, respectively. The scale factors are computed based on the size of the bounding boxes of $\boldsymbol{T}$ and $\boldsymbol{S}$. Due to the articulated motion, the poses of $\boldsymbol{T}$ and $\boldsymbol{S}$ are different. Thus, the initial alignment procedure is performed, separately, for the forearm and upper arm, in this study.


\subsubsection{Non-Rigid Registration and Scanning Trajectory Determination}
\par
After initial alignments, the registration between the surface point cloud $\boldsymbol{T}$ and $\boldsymbol{S}$ is carried out to transfer the computed scanning trajectory from the atlas to the real environment. Considering patient-specific parameters and inevitable articulated motion around joints, a non-rigid registration approach proposed by Yao~\emph{et al.}~\cite{yao2020quasi} is used. To reduce the number of variables and improve the time efficiency, a deformation graph is first reconstructed by filtering the vertices of $\boldsymbol{S}$ using geodesic distance. In the resulting graph, the geodesic distance between any two consecutive vertices is larger than a preset threshold. Then the non-rigid registration issue is considered an optimization problem. To properly characterize the non-rigid registration problem, an optimization function $L_{nr}$ is built by considering the consistency between individual vertices alignment from source and target point clouds $L_{ali}(X)$, the consistency of local neighboring vertices \revision{$L_{reg}(X)$} and the rigidity of local surface $L_{rot}(X)$ as Eq.~(\ref{Eq_non_rigid_registration}).

\begin{equation}\label{Eq_non_rigid_registration}
\min_M L_{nr} = L_{ali}(M)+ \alpha_1 L_{reg}(M)+ \alpha_2 L_{rot}(M)
\end{equation}
where $M$ is the set of the \revision{computed transformation matrix} mapping the points from $\boldsymbol{S}$ to $\boldsymbol{T}$. $\alpha_1$ and $\alpha_2$ are the weight coefficients. For more details related to the solving of the optimization process, please refer to~\cite{yao2020quasi}.  

\par
\begin{figure}[ht!]
\centering
\includegraphics[width=0.45\textwidth]{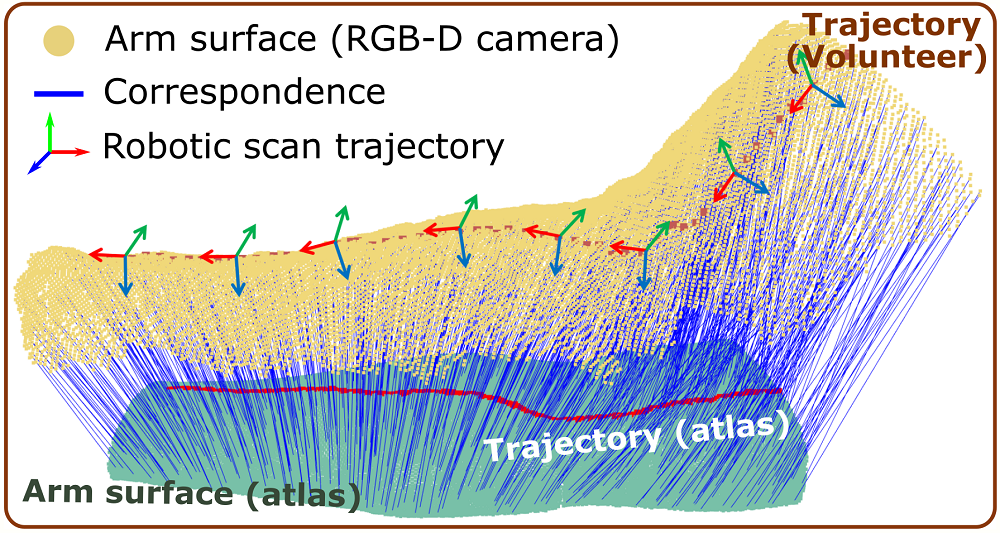}
\caption{A representative non-rigid registration result. The pose of arm is different with the one used in atlas due to the articulated motion.
}
\label{Fig_non_rigid}
\end{figure}

A representative non-rigid registration result has been depicted in Fig.~\ref{Fig_non_rigid}. Although the pose of the real human arm is different from the one used in the atlas due to the articulated motion, reasonable correspondences between the points among $\boldsymbol{S}$ and $\boldsymbol{T}$ are achieved (see blue lines). Based on the optimized transform matrix $M$, each point on the planned trajectory can be projected to the real human arm as shown in Fig.~\ref{Fig_non_rigid}. \revision{To determine the probe pose, the long axis of the probe is placed to be perpendicular to the scan direction, and the probe centerline is aligned to the normal direction of the extracted surface point cloud from the camera.} To further generate executable scanning trajectory for the robotic manipulator, the hand-eye calibrations method described in~\cite{jiang2021motion} is employed in this study. \revision{We first manually obtain the coordinate representations of the same intersection points ($>4$) on two checkerboards placed at different heights in the robotic base frame and camera frame. Then, a rigid transformation between the camera frame and the robotic base frame can be optimized. 
}


\subsection{Object-Specific Centering}
\par
Due to the errors caused by the non-rigid registration and hand-eye calibration, the object may not be properly visualized or even out of the US imaging field. To alleviate the negative impact caused by such errors, an object-specific centering method is developed based on the position of the segmented vessels. Based on the moments $M_{00}$ and $M_{10}$ for the binary masks obtained from the network, the horizontal position of the vessel $v_x$ in the image is computed as $v_x=M_{10}/M_{00}$.

\par
If the computed centroid of the vessel of interest deviates from the vertical centerline of the US view, a relative movement in the robotic base frame is computed using $\Delta P = {^{b}_{I}T}~[(W_p/2-v_x), 0, 0, 0]^T$, where $^{b}_{I}T$ is the homogeneous transformation matrix mapping the US image pixel positions to the robotic base frame and $W_p$ is the pixel-wise width of US images. To guarantee the smoothness of the scanning trajectory, the computed $\Delta P$ is extrapolated for the remaining points as Eq.~(\ref{Eq_compensation_extrapolat}).

\begin{equation}\label{Eq_compensation_extrapolat}
P'_{i+k}=P_{i+k}+\Delta P_i\sigma^{k}
\end{equation}
where $i$ is the iterator representing the centering compensation procedure happens, $k= 1,2,3 ...$ are the points after $i$-th point, and $0.5<\sigma<1$ is the weight used to reduce the effectiveness of the extrapolated movement as distance increases.



\section{Result}
\subsection{Experimental Setup}
\par
\begin{figure}[ht!]
\centering
\includegraphics[width=0.40\textwidth]{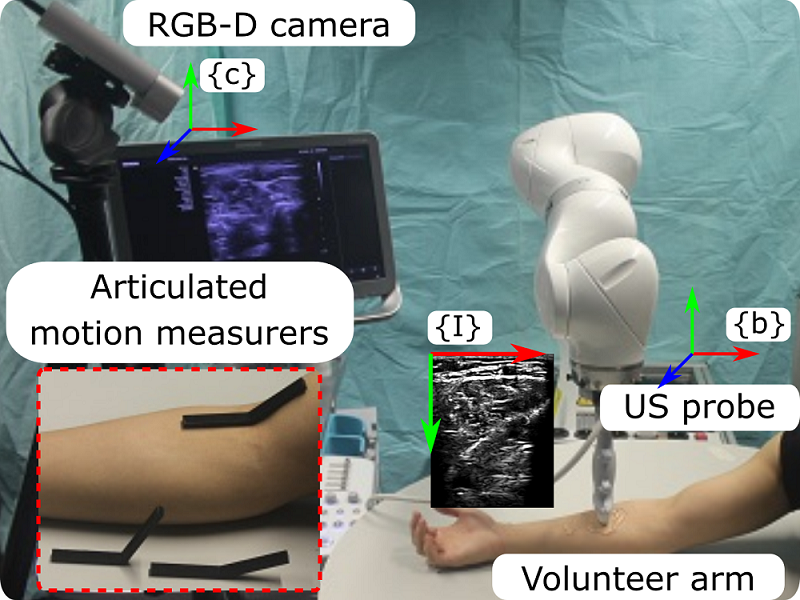}
\caption{Experimental setup. \revision{$\{c\}$, $\{b\}$ and $\{I\}$ are the coordinate frames of the camera, robotic base and US images, respectively.}
}
\label{Fig_experimental_setup}
\end{figure}

The overall experimental setup is presented in Fig.~\ref{Fig_experimental_setup}. A linear probe is attached on the end-effector of the robotic arm (LBR iiwa 14 R820, KUKA GmbH, Germany). The robot is controlled using a self-developed Robot Operating System (ROS) interface~\cite{hennersperger2016towards}. To guarantee the real-time performance, the robot is controlled at $100~Hz$. To validate the correctness of the generated scanning trajectory, six volunteers are employed. The MRI image obtained from one volunteer is used as the atlas template. To further consider articulated motion, three measures with fixed angles ($160^{\circ}$, $140^{\circ}$ and $120^{\circ}$) are 3D-printed as shown in Fig.~\ref{Fig_experimental_setup}. 

\par
\revision{To perform the scan along the planned trajectory with constant pressure, a compliant controller using the built-in joint torque sensors was deployed. A given force is automatically maintained in the direction of the probe centerline by a 1-DoF compliant controller, while the other 5-DoF motion is controlled to accurately follow the planned sweep trajectory. These two types of controllers are fused by assigning varied stiffness values. High stiffness (translational DoF: $2000~N/m$ and rotational DoF: $200~Nm/m$) are assigned to the DoFs controlled in position mode, while the stiffness of the compliant DoF is set between $[125, 500]~N/m$~\cite{hennersperger2016towards}. More details about the controller can refer~\cite{jiang2021autonomous,jiang2020automaticTIE}.
}

\subsection{Performance of the Proposed Network}
\par
To properly train the proposed OF-UNet, the classic metric dice coefficient $d_c = \frac{2~|G\cap S|}{|G|+|S|}$ is employed in this study, where $G$ is the set of the pixels classified as the object in the ground truth and $S$ is the set of pixels classified as the object from OF-UNet. $d_c$ is often used to measure the similarity between the segmentation result and the ground truth. The loss function is defined as $\mathfrak{L}_d = 1-d_c$ and the ADAM optimizer is used to minimize $\mathfrak{L}_d$. The learning rate was set to $0.001$ at the beginning and decreased by a factor of ten when the loss changes were less than $0.0001$ for ten successive steps. The OF-Unet model is trained on a workstation (GPU: Titan X GPU; CPU: i7 7700K). The performance of the proposed OF-Unet on in-vivo radial arteries obtained from two unseen volunteers has been summarized in TABLE~\ref{Table_segmentation_result}. The results are computed on $250$ images of each volunteer. 


\begin{table}[!ht]
\centering
\caption{Segmentation Results on Unseen Volunteers (Mean$\pm$SD)}
\label{Table_segmentation_result}
\begin{tabular}{ccccc}
\noalign{\hrule height 1.2pt}
Data                        & Methods & dice       & precision   & recall    \\
\noalign{\hrule height 1.0pt}
\multirow{2}{*}{Volunteer1} & UNet    &  $0.80\pm0.15$   &  $0.76\pm0.17$     & $0.85\pm0.15$         \\
                            & \textbf{OF-UNet} &  $0.84\pm0.06$   &  $0.82\pm0.06$     & $0.88\pm0.09$  \\
\noalign{\hrule height 0.8pt}
\multirow{2}{*}{Volunteer2} & UNet    &  $0.79\pm0.11$   &  $0.72\pm0.15$     & $0.90\pm0.07$         \\ 
                            & \textbf{OF-UNet} &  $0.86\pm0.04$   &  $0.78\pm0.05$     & $0.95\pm0.04$ \\ 
\noalign{\hrule height 1.2pt}
\end{tabular}
\end{table}



\par
It can be seen from TABLE~\ref{Table_segmentation_result} that, better performance is achieved by the proposed OF-Unet on the unseen volunteers in three different metrics (dice coefficient, precision, and recall). The results computed on different volunteers also demonstrated good consistency. In addition, it is also noteworthy that the standard deviation of the proposed OF-Unet is smaller than the ones of the standard U-Net. This means that combining the optical flow information can result in more robust segmentation results. This characteristic is pretty useful for the task requiring segmentation and tracking simultaneously, i.e., screening of a blood vessel.

\subsection{Cross-Validation of Trajectory Generation}
\par
\begin{figure}[ht!]
\centering
\includegraphics[width=0.47\textwidth]{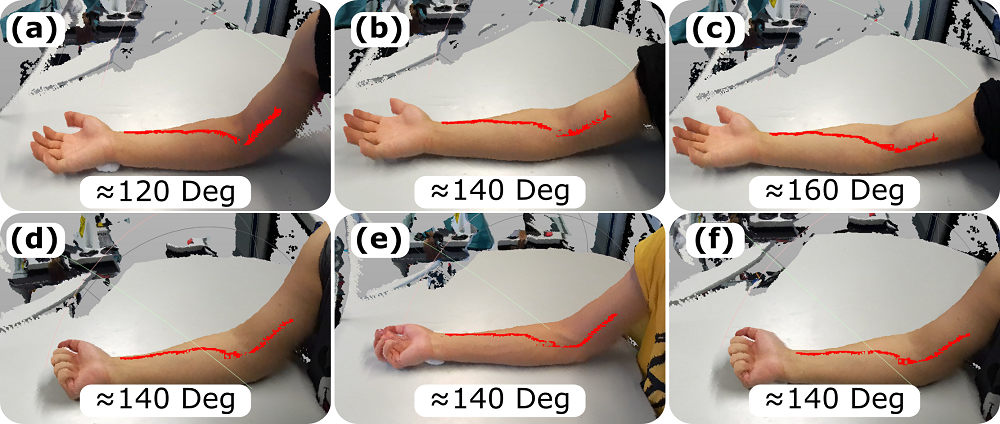}
\caption{Representative trajectories generated on different volunteers' arm surfaces in RGB point cloud view. (a), (b) and (c) are the generated trajectories on the volunteer whose MRI data is used as atlas template when the elbow joint angles is around $120^{\circ}$, $140^{\circ}$ and $160^{\circ}$, respectively. (d), (e) and (f) are the results on three different volunteers without exact MRI data when elbow joint angles is around $140^{\circ}$. The red lines are the generated trajectories.  
}
\label{Fig_trajectory_generation}
\end{figure}

To validate whether the proposed approach can properly generate the scan trajectory for scanning when the arm pose is different from the one in MRI data, experiments are carried out on six volunteers. The MRI template is obtained from one of the volunteers when the arm is horizontally placed, namely articulated joint angle is around $180^{\circ}$. All volunteers are asked to position their arm in three different poses (the articulated joint angle around $120^{\circ}$, $140^{\circ}$, and $160^{\circ}$). A few representative results have been presented in Fig.~\ref{Fig_trajectory_generation}. Fig.~\ref{Fig_trajectory_generation} (a), (b) and (c)
are the results from the same volunteer with different poses. Reasonable trajectories, similar to the distribution of radial arteries, are generated in all cases. In addition, we further validate the performance of the proposed method on other volunteers without their own MRI data. Three representative results obtained from three volunteers when their arms are positioned in $140^{\circ}$ are depicted in Fig.~\ref{Fig_trajectory_generation} (d), (e) and (f). The results demonstrate that the proposed non-rigid approach can effectively and robustly generate proper trajectories for different patients, even in presence of articulated motion.


\par



\subsection{Quantitative Validation of the Vascular Geometry}
\par
\begin{figure}[ht!]
\centering
\includegraphics[width=0.47\textwidth]{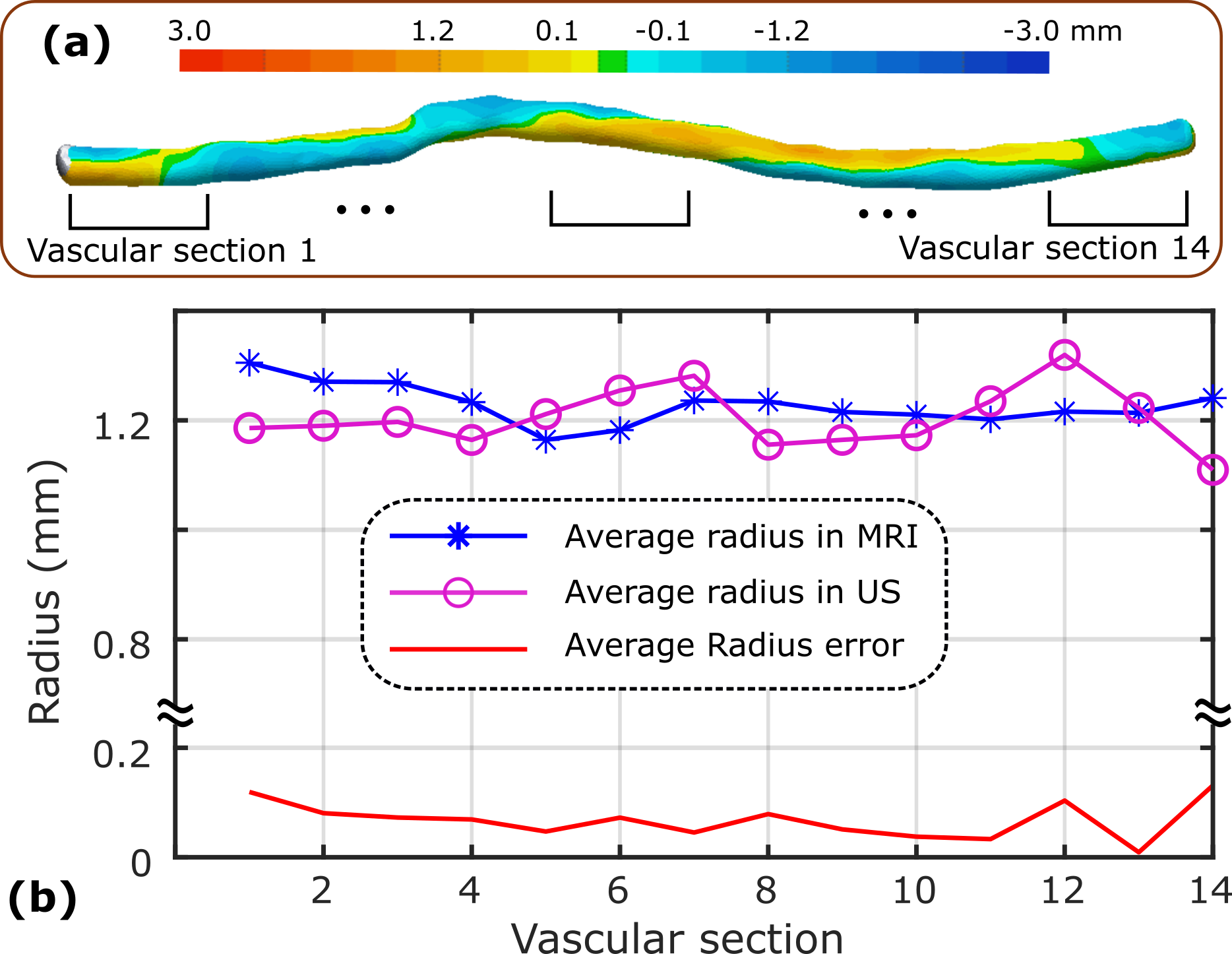}
\caption{Quantitative results of robotic US scan in terms of radius. (a) 3D heatmap of the geometry difference between the obtained US image and MRI. (b) Average radius of different vascular sections.}
\label{Fig_vessel_overlap}
\end{figure}

To quantitatively evaluate the final results, the reconstructed 3D images of the radial arterial obtained from robotic US scanning are registered to the 3D vessel ground truth obtained from MRI data. The differences between the two 3D images have been intuitively visualized in Fig.~\ref{Fig_vessel_overlap}~(a). It can be seen that the error of most individual points is less than $0.4~mm$. However, the geometrical error of individual points is severely affected by the alignment, particularly for the small arteries like radial ($1.18\pm0.2~mm$~\cite{beniwal2014size}). To further evaluate the performance of the radius estimation, the average radius of the 3D images obtained from MRI and US are $1.2\pm0.04~mm$ and $1.2\pm0.05~mm$, respectively, for a radial artery segmentation (length: $70~mm$) on the same volunteer. The results obtained from the autonomous US scanning are comparable to the results obtained from MRI data in terms of average radius. 

\par
To further investigate local variations, the radial artery segmentation is further equally divided into $14$ sub-segments ($\approx5~mm$ for each) as in Fig.~\ref{Fig_vessel_overlap}~(a). The average radius for each sub vascular section has been presented in Fig.~\ref{Fig_vessel_overlap}~(b). The mean error between MRI and and US is only $0.06~mm$ and all error for the $14$ sub sections are less than $0.13~mm$.  




\par

\section{Discussion}
\par
\revision{
We present a novel vision-based RUSS that can robustly generate an optimal scanning trajectory on different patients' skin surfaces for autonomous US scanning of limb arteries, even in presence of articulated motion. To the best of the authors' knowledge, this is the first work considering the inevitable joint articulated motion for autonomous RUSS. The main factors that limit the achievement of repeatable and accurate US scans are patients' motion (both rigid and articulated motion) and image deformations caused by varying pressure during scans. Thereby, by further integrating the proposed method dealing with articulated motions to the developed motion-aware~\cite{jiang2021motion} (rigid motion) and deformation-aware~\cite{jiang2021deformation} work, we can improve the clinical acceptance of RUSS. }

\revision{
The scanning trajectory generation approach and the proposed OF-UNet both perform boldly on different volunteers in our experiments. However, there are still some limitations that need to be discussed. First, due to the limitation of robotic working space, the trajectory only covers one-third of the upper arm and full forearm until the wrist joint as Figs.~\ref{Fig_arm_segmentation} and \ref{Fig_trajectory_generation}. The full scan from the shoulder joint to the wrist hasn't been done yet. However, the proposed method can be directly used to compute a full scan path starting from the shoulder joint. 
In addition, to obtain a smooth scanning trajectory, the orientation optimization approach developed in~\cite{jiang2021motion} can be further employed to eliminate the out-of-plane motion during scans. These motions are caused by the estimation error of local normal direction at contact point using RGBD data.  
Finally, the proposed OF-UNet is only tested on healthy volunteers. Future work will further test the proposed approach in clinical on real patients of different ages, weights, and heights should be investigated with the help of clinicians.
}

\section{Conclusion}
\par
This work presents a vision-based approach enabling autonomous robotic US scans for limb arteries by further considering the articulated motion. The method demonstrates the potential of the automatic scanning framework, allowing efficient and accurate robotic 3D US acquisitions based on generic atlas with pre-planned trajectories. A non-rigid registration between the surface point cloud obtained from the real environment and the atlas is carried out to properly map the planned trajectory to an executable path for RUSS. To evaluate the robustness and effectiveness, trajectories were properly generated on six volunteers with different arm poses (see Fig.~\ref{Fig_trajectory_generation}). Besides, to effectively segment and track the target arteries (e.g., radial arteries), a novel OF-UNet is proposed to fully leverage the spatial continuity among the consecutive US frames to facilitate the segmentation. The proposed OF-UNet outperforms the standard U-Net in terms of different metrics on two unseen volunteers' radial images (see TABLE~\ref{Table_segmentation_result}). The experimental results demonstrate that the geometry measurements obtained from robotic US images ($1.2\pm0.05~mm$) are comparable to the values obtained from MRI data ($1.2\pm0.04~mm$). By further considering the challenging factors in real scenarios, i.e., the articulated motion, we consider the proposed RUSS can prove the feasibility of autonomous robotic US scans and hopefully improve the clinical acceptance of such systems. 

%% file: root.bbl
\begin{thebibliography}{10}
\providecommand{\url}[1]{#1}
\csname url@samestyle\endcsname
\providecommand{\newblock}{\relax}
\providecommand{\bibinfo}[2]{#2}
\providecommand{\BIBentrySTDinterwordspacing}{\spaceskip=0pt\relax}
\providecommand{\BIBentryALTinterwordstretchfactor}{4}
\providecommand{\BIBentryALTinterwordspacing}{\spaceskip=\fontdimen2\font plus
\BIBentryALTinterwordstretchfactor\fontdimen3\font minus
  \fontdimen4\font\relax}
\providecommand{\BIBforeignlanguage}[2]{{%
\expandafter\ifx\csname l@#1\endcsname\relax
\typeout{** WARNING: IEEEtran.bst: No hyphenation pattern has been}%
\typeout{** loaded for the language `#1'. Using the pattern for}%
\typeout{** the default language instead.}%
\else
\language=\csname l@#1\endcsname
\fi
#2}}
\providecommand{\BIBdecl}{\relax}
\BIBdecl

\bibitem{favaretto2007analysis}
E.~Favaretto, C.~Pili, and et~al., ``Analysis of agreement between duplex
  ultrasound scanning and arteriography in patients with lower limb artery
  disease,'' \emph{J. Cardiovasc. Med.}, vol.~8, no.~5, pp. 337--341, 2007.

\bibitem{merouche2015robotic}
S.~Merouche, L.~Allard, E.~Montagnon, G.~Soulez, P.~Bigras, and G.~Cloutier,
  ``A robotic ultrasound scanner for automatic vessel tracking and
  three-dimensional reconstruction of b-mode images,'' \emph{IEEE Trans.
  Ultrason. Ferroelectr. Freq. Control}, vol.~63, no.~1, pp. 35--46, 2015.

\bibitem{morgan2018versatile}
M.~R. Morgan, J.~S. Broder, and et~al., ``Versatile low-cost volumetric 3-d
  ultrasound platform for existing clinical 2-d systems,'' \emph{IEEE Trans.
  Med. Imaging}, vol.~37, no.~10, pp. 2248--2256, 2018.

\bibitem{jiang2021deformation}
Z.~Jiang, Y.~Zhou, Y.~Bi, M.~Zhou, T.~Wendler, and N.~Navab,
  ``Deformation-aware robotic 3d ultrasound,'' \emph{IEEE Robot. Autom. Lett.},
  vol.~6, no.~4, pp. 7675--7682, 2021.

\bibitem{gilbertson2015force}
M.~W. Gilbertson and B.~W. Anthony, ``Force and position control system for
  freehand ultrasound,'' \emph{IEEE Trans. Robot.}, vol.~31, no.~4, pp.
  835--849, 2015.

\bibitem{pierrot1999hippocrate}
F.~Pierrot, E.~Dombre, E.~D{\'e}goulange, L.~Urbain, P.~Caron, S.~Boudet,
  J.~Gari{\'e}py, and J.-L. M{\'e}gnien, ``Hippocrate: A safe robot arm for
  medical applications with force feedback,'' \emph{Med. Image Anal.}, vol.~3,
  no.~3, pp. 285--300, 1999.

\bibitem{jiang2020automaticTIE}
Z.~Jiang, M.~Grimm, M.~Zhou, Y.~Hu, J.~Esteban, and N.~Navab, ``Automatic
  force-based probe positioning for precise robotic ultrasound acquisition,''
  \emph{IEEE Transactions on Industrial Electronics}, vol.~68, no.~11, pp.
  11\,200--11\,211, 2020.

\bibitem{jiang2020automatic}
Z.~Jiang, M.~Grimm, M.~Zhou, J.~Esteban, W.~Simson, G.~Zahnd, and N.~Navab,
  ``Automatic normal positioning of robotic ultrasound probe based only on
  confidence map optimization and force measurement,'' \emph{IEEE Robot. Autom.
  Lett.}, vol.~5, no.~2, pp. 1342--1349, 2020.

\bibitem{ihnatsenka2010ultrasound}
B.~Ihnatsenka and A.~P. Boezaart, ``Ultrasound: Basic understanding and
  learning the language,'' \emph{Int. J. Shoulder Surg.}, vol.~4, no.~3, p.~55,
  2010.

\bibitem{hennersperger2016towards}
C.~Hennersperger, B.~Fuerst, and et~al., ``Towards {MRI}-based autonomous
  robotic us acquisitions: a first feasibility study,'' \emph{IEEE Trans. Med.
  Imaging}, vol.~36, no.~2, pp. 538--548, 2016.

\bibitem{virga2016automatic}
S.~Virga, O.~Zettinig, M.~Esposito, K.~Pfister, B.~Frisch, T.~Neff, N.~Navab,
  and C.~Hennersperger, ``Automatic force-compliant robotic ultrasound
  screening of abdominal aortic aneurysms,'' in \emph{Proc. IEEE/RSJ Int. Conf.
  Intell. Robot. Syst. (IROS)}.\hskip 1em plus 0.5em minus 0.4em\relax IEEE,
  2016, pp. 508--513.

\bibitem{jiang2021motion}
Z.~Jiang, H.~Wang, and et~al., ``Motion-aware robotic 3d ultrasound,'' in
  \emph{Proc. IEEE Int. Conf. Robot. Automat. (ICRA)}.\hskip 1em plus 0.5em
  minus 0.4em\relax IEEE, 2021.

\bibitem{mishra2018ultrasound}
D.~Mishra, S.~Chaudhury, M.~Sarkar, and A.~S. Soin, ``Ultrasound image
  segmentation: a deeply supervised network with attention to boundaries,''
  \emph{IEEE Trans. Biomed. Eng.}, vol.~66, no.~6, pp. 1637--1648, 2018.

\bibitem{frangi1998multiscale}
A.~F. Frangi, W.~J. Niessen, K.~L. Vincken, and M.~A. Viergever, ``Multiscale
  vessel enhancement filtering,'' in \emph{International conference on medical
  image computing and computer-assisted intervention}.\hskip 1em plus 0.5em
  minus 0.4em\relax Springer, 1998, pp. 130--137.

\bibitem{karami2018adaptive}
E.~Karami, M.~S. Shehata, and A.~Smith, ``Adaptive polar active contour for
  segmentation and tracking in ultrasound videos,'' \emph{IEEE Trans. Circuits
  Syst. Video Technol.}, vol.~29, no.~4, pp. 1209--1222, 2018.

\bibitem{smistad2015real}
E.~Smistad and F.~Lindseth, ``Real-time automatic artery segmentation,
  reconstruction and registration for ultrasound-guided regional anaesthesia of
  the femoral nerve,'' \emph{IEEE Trans. Med. Imaging}, vol.~35, no.~3, pp.
  752--761, 2015.

\bibitem{abolmaesumi2002image}
P.~Abolmaesumi, S.~E. Salcudean, W.-H. Zhu, M.~R. Sirouspour, and S.~P. DiMaio,
  ``Image-guided control of a robot for medical ultrasound,'' \emph{IEEE Trans.
  Robot. Autom.}, vol.~18, no.~1, pp. 11--23, 2002.

\bibitem{litjens2017survey}
G.~Litjens, T.~Kooi, B.~E. Bejnordi, A.~A.~A. Setio, F.~Ciompi, M.~Ghafoorian,
  J.~A. Van Der~Laak, B.~Van~Ginneken, and C.~I. S{\'a}nchez, ``A survey on
  deep learning in medical image analysis,'' \emph{Medical image analysis},
  vol.~42, pp. 60--88, 2017.

\bibitem{ronneberger2015u}
O.~Ronneberger, P.~Fischer, and T.~Brox, ``U-net: Convolutional networks for
  biomedical image segmentation,'' in \emph{International Conference on Medical
  image computing and computer-assisted intervention}.\hskip 1em plus 0.5em
  minus 0.4em\relax Springer, 2015, pp. 234--241.

\bibitem{jiang2021autonomous}
Z.~Jiang, Z.~Li, M.~Grimm, M.~Zhou, M.~Esposito, W.~Wein, W.~Stechele,
  T.~Wendler, and N.~Navab, ``Autonomous robotic screening of tubular
  structures based only on real-time ultrasound imaging feedback,'' \emph{IEEE
  Transactions on Industrial Electronics}, 2021.

\bibitem{chen2020deep}
A.~I. Chen, M.~L. Balter, T.~J. Maguire, and M.~L. Yarmush, ``Deep learning
  robotic guidance for autonomous vascular access,'' \emph{Nature Machine
  Intelligence}, vol.~2, no.~2, pp. 104--115, 2020.

\bibitem{jiang2021automatic_baichuan}
B.~Jiang, A.~Chen, S.~Bharat, and M.~Zheng, ``Automatic ultrasound vessel
  segmentation with deep spatiotemporal context learning,'' in
  \emph{International Workshop on Advances in Simplifying Medical
  Ultrasound}.\hskip 1em plus 0.5em minus 0.4em\relax Springer, 2021, pp.
  3--13.

\bibitem{ma2021autonomous}
X.~Ma, Z.~Zhang, and H.~K. Zhang, ``Autonomous scanning target localization for
  robotic lung ultrasound imaging,'' in \emph{Proc. IEEE/RSJ Int. Conf. Intell.
  Robot. Syst. (IROS))}.\hskip 1em plus 0.5em minus 0.4em\relax IEEE, 2021, pp.
  9467--9474.

\bibitem{huang2021towards}
Y.~Huang, W.~Xiao, C.~Wang, H.~Liu, R.~Huang, and Z.~Sun, ``Towards fully
  autonomous ultrasound scanning robot with imitation learning based on
  clinical protocols,'' \emph{IEEE Robot. Autom. Lett.}, vol.~6, no.~2, pp.
  3671--3678, 2021.

\bibitem{welleweerd2021out}
M.~Welleweerd, A.~de~Groot, V.~Groenhuis, F.~Siepel, and S.~Stramigioli,
  ``Out-of-plane corrections for autonomous robotic breast ultrasound
  acquisitions,'' in \emph{Proc. IEEE Int. Conf. Robot. Automat. (ICRA)}.\hskip
  1em plus 0.5em minus 0.4em\relax IEEE, 2021, pp. 12\,515--12\,521.

\bibitem{bi2022vesnet}
Y.~Bi and et~al., ``Ves{N}et-{RL}: Simulation-based reinforcement learning for
  real-world us probe navigation,'' \emph{IEEE Robot. Autom. Lett.}, pp. 1--1,
  2022.

\bibitem{beniwal2014size}
S.~Beniwal, K.~Bhargava, and S.~K. Kausik, ``Size of distal radial and distal
  ulnar arteries in adults of southern rajasthan and their implications for
  percutaneous coronary interventions,'' \emph{Indian heart journal}, vol.~66,
  no.~5, pp. 506--509, 2014.

\bibitem{ilg2017flownet}
E.~Ilg, N.~Mayer, T.~Saikia, M.~Keuper, A.~Dosovitskiy, and T.~Brox, ``Flownet
  2.0: Evolution of optical flow estimation with deep networks,'' in
  \emph{Proceedings of the IEEE conference on computer vision and pattern
  recognition}, 2017, pp. 2462--2470.

\bibitem{yao2020quasi}
Y.~Yao, B.~Deng, W.~Xu, and J.~Zhang, ``Quasi-newton solver for robust
  non-rigid registration,'' in \emph{Proceedings of the IEEE/CVF conference on
  computer vision and pattern recognition}, 2020, pp. 7600--7609.

\end{thebibliography}
